\documentclass[10pt,letterpaper]{article}
\usepackage[top=0.85in,footskip=0.75in,marginparwidth=2in]{geometry}

\usepackage{nameref,hyperref}

\usepackage{booktabs}

\usepackage[aboveskip=1pt,labelfont=bf,labelsep=period,singlelinecheck=off]{caption}

\makeatletter
\renewcommand{\@biblabel}[1]{\quad#1.}
\makeatother

\usepackage{amsmath}
\usepackage{graphicx,psfrag,epsf}
\usepackage{natbib}

\begin{document}
	\vspace*{0.35in}
	
	\begin{flushleft}
		{\Large
			\textbf\newline{A Comparison of Resampling and Recursive Partitioning Methods in Random Forest for Estimating the Asymptotic Variance Using the Infinitesimal Jackknife}
		}
		\newline
		\\
		Cole Brokamp\textsuperscript{1,*},
		MB Rao\textsuperscript{2},
		Patrick Ryan\textsuperscript{1},
		Roman Jandarov\textsuperscript{2},
		\\
		\bigskip
		\textsuperscript{1} Division of Biostatistics and Epidemiology, Cincinnati Children's Hospital Medical Center
		\\
		\textsuperscript{2} Department of Environmental Health, University of Cincinnati
		\\
		\textsuperscript{*} cole.brokamp@cchmc.org
		
	\end{flushleft}
	
	\section*{Abstract}
	The infinitesimal jackknife (IJ) has recently been applied to the random forest to estimate its prediction variance. These theorems were verified under a traditional random forest framework which uses classification and regression trees (CART) and bootstrap resampling.  However, random forests using conditional inference (CI) trees and subsampling have been found to be not prone to variable selection bias. Here, we conduct simulation experiments using a novel approach to explore the applicability of the IJ to random forests using variations on the resampling method and base learner. Test data points were simulated and each trained using random forest on one hundred simulated training data sets using different combinations of resampling and base learners. Using CI trees instead of traditional CART trees as well as using subsampling instead of bootstrap sampling resulted in a much more accurate estimation of prediction variance when using the IJ. The random forest variations here have been incorporated into an open source software package for the R programming language.

\section{Introduction}

\subsection{Random Forest}
\label{section:rf_introduction}

Although random forests are commonly used in machine learning, they still remain underused for statistical inference because of a lack understanding of their statistical properties. A recent study found random forest to be the most accurate classification algorithm among 179 classifiers, based on 121 different data sets \citep{fernandez2014we}. Although proven to be more accurate, researchers are sometime hesitant to implement random forests because they do not have parameters with a clear interpretation like regression coefficients from parametric models.

Random forest is an ensemble learning method that begins with bagging (the bootstrapped aggregation of regression tree predictions) in order to reduce the variance of the prediction function.  Here, a bootstrapped sample refers to new sample taken from the original sample with replacement. Regression trees are low in bias, and although they have high variance bagging stabilizes the predictions leading to lower variance than using one tree \citep{Hastie}. The bagging procedure was modified by \cite{Brieman} to use a bootstrap sample for each tree and to also select a random subset of predictors for testing at each split point in each tree.  This de-correlates individual trees, further reducing the ensemble prediction variance.

The specific algorithm for random forest as used for regression is as follows:

\begin{enumerate}
	\def\labelenumi{\arabic{enumi}.}
	\itemsep1pt\parskip0pt\parsep0pt
	\item
	For $b=1$ to $B$ total trees:
	
	\begin{itemize}
		\itemsep1pt\parskip0pt\parsep0pt
		\item
		Draw a bootstrap sample from the training data
		\item
		Grow tree $T_b$ by repeating the following steps for each terminal
		node of the tree until the desired node size is reached:
		
		\begin{itemize}
			\itemsep1pt\parskip0pt\parsep0pt
			\item
			Randomly select $m_{try}$ of the total $p$ variables
			\item
			Pick the best variable and split-point from the $m_{try}$ variables
			based on the best reduction in the sum of the squared errors of
			the predictions
			\item
			Split the node into two daughter nodes
		\end{itemize}
	\end{itemize}
	\item
	Output the total ensemble of all trees.
	\item
	To predict at a new point $x$, average the prediction of all trees:
	$\hat{f}(x)=\frac{1}{B}\sum\limits_{b=1}^{B}{T_b(x)}$
\end{enumerate}

The performance of random forests can be tuned using two parameters, $m_{try}$ and $B$.  $B$ is the total number of trees and is set to 500 by default in the \texttt{randomForest} package within \texttt{R}.  The number of trees should be large enough so that the error rate is stabilized.  Since the random forest is grown one tree at a time, the error rate can be plotted as a function of the number of trees to visually ensure that enough trees are being used. $m_{try}$ usually has more effect on the ensemble accuracy and is set to $max\{floor(\frac{1}{3}p),1\}$ as the default in the \texttt{randomForest} package within \texttt{R}.  Variations in $m_{try}$ can be auditioned and the value producing the lowest error can be used in the final random forest model.

\subsection{Estimating the Variance of Bagged Tree Predictions Using the Jackknife}

Bootstrap sampling, subsampling, and the jackknife all rely on estimating the variance of a statistic by using the variability between resamples rather than using statistical distributions. The ordinary jackknife is a resampling method useful for estimating the variance or bias of a statistic.  The jackknife estimate of a statistic can be found by repeatedly calculating the statistic, each time leaving one observation from the sample out and averaging all estimates.  The variance of the estimate can be found by calculating the variance of the jackknifed estimates:

\begin{equation}
\hat{V}_{J} = \frac{n-1}{n}\sum\limits_{i=1}^{n} \left(\hat{\theta}_{(-i)} - \hat{\theta}_{(\cdot)}\right)^2
\end{equation}

where $n$ is the total sample size, $\hat{\theta}_{(-i)}$ is the statistic estimated without using the $i^{th}$ observation, and $\hat{\theta}_{(\cdot)}$ is the average of all jackknife estimates.

The ordinary jackknife is extended for use with bagging by applying it to the bootstrap distribution \cite{Efron}.  Instead of leaving out one observation at a time, the existing bootstrap samples are used and the statistic is calculated based on all resamples which do not use the $i^{th}$ observation:

\begin{equation}
\hat{V}_{JB} = \frac{n-1}{n}\sum\limits_{i=1}^{n} \left(\bar{t^*}_{(-i)}(x) - \bar{t^*}_{(\cdot)}(x)\right)^2
\end{equation}

where $\bar{t^*}_{(-i)}(x)$ is the average of $t^*(x)$ over all bootstrap samples not containing the $i^{th}$ example and $\bar{t^*}_{(\cdot)}(x)$ is the mean of all $\bar{t^*}_{(i)}(x)$. 

\subsection{Infinitesimal Jackknife}

As opposed to the jackknife and the jackknife after bootstrap, where the behavior of a statistic is studied after removing one or more observations at a time, the IJ looks at the behavior of a statistic after down-weighting each observation by an infinitesimal amount \citep{jaeckel1972infinitesimal}. Applied to a bagged predictor, the non-parametric delta-method estimate of variance for an ideal smoothed bootstrap statistic is \citep{Efron}:

\begin{equation}
    \hat{V}_{IJ}=\sum_{j=1}^{n}\text{cov}_j
\end{equation}

where $cov_j$ is taken with respect to the resampling distribution. \cite{Wager} have recently extended this idea by applying the IJ to random forest predictions.  Based on using subsamples rather than bootstrap samples, they have shown that the variance of random forest predictions can be consistently estimated.  Here the IJ variance estimator is applied to the resampling distribution for a new prediction point:
\begin{equation}
\label{eq:IJ}
\hat{V}_{IJ} = \sum\limits_{i=1}^{n} Cov_*\left[T(x; Z^*_1,...,Z^*_n),N^*_i\right]
\end{equation}
where $T(x; Z^*_1,...,Z^*_n)$ is the prediction of the tree $T$ for the test point $x$ based on the subsample $Z^*_1,...,Z^*_n$ and $N^*_i$ is the number of times $Z_i$ appears in the subsample. Furthermore, random forest predictions are asymptotically normal given that the underlying trees are based on subsampling and that the subsample size $s$ scales as $s(n)/n=o(log(n)^{-p})$, where $n$ the is number of training examples and $p$ is the number of features \cite{Wager2}.  

Because $\hat{V}_{IJ}$ is calculated in practice with a finite number of trees $B$, it is inherently associated with Monte Carlo error. Although this error can be decreased by using a large $B$, a correction has been suggested \cite{Wager}:

\begin{equation}
\label{eq:IJB}
\hat{V}_{IJ}^B = \sum\limits_{i=1}^{n}C_i^2 - \frac{s(n-s)}{n}\frac{\hat{v}}{B}
\end{equation}

where $C_i = \frac{1}{B}\sum\limits_{b=1}^{B}(N^*_{bi} - s/n)(T^*_b - \bar{T}^*)$ and $\hat{v} = \frac{1}{B}\sum\limits_{b=1}^{B}(T^*_b - \bar{T}^*)^2$. This is essentially a Monte Carlo estimate of Equation \ref{eq:IJ} with a bias correction subtracted off. These estimates are asymptotically normal given a few key conditions, one of which is that the underlying trees are honest. Simulation experiments using sub bagged random forests have shown that these variance estimates are biased \cite{Wager2}, but the implementation of honest trees within a sub bagged tree ensemble and its resulting prediction variance has not been studied.

\subsection{Using Honest Trees in Random Forests}

\cite{athey2016recursive} (and \cite{Wager2} within the context of random forests) define an honest tree as one in which the distribution of the predicted outcome, conditional on the explanatory variables, does not depend on the training labels. The most popular recursive partitioning algorithm and the one used in the random forest algorithm is CART \citep{Breiman}. In the case of regression, this algorithm performs a search for the best possible split over all split points of all variables by minimizing the sum of squared errors between the predicted and actual values. CART trees are not honest because they use the same training data to both choose the tree splits and to make predictions. In contrast, 
conditional inference (CI) trees \citep{hothorn2006unbiased} are trees that are honest because they use outcomes to make predictions but use another method to find split points. CI trees implement a test statistic, like the Spearman correlation coefficient, student's t test, or F statistic from ANOVA to pick the predictor that is most associated with the outcome based on the smallest p-value. P-values are generated using a permutation test framework first laid out by \cite{Strasser} in which the distribution of the test statistic under the null hypothesis is obtained by calculating all possible values of the test statistic under rearrangements of the labels on the observed data points. To find the best split point, the standardized test statistic is then maximized. \cite{Strobl} showed that implementing CI trees within a random forest framework alleviates variable selection bias, which favors splitting on variables with more levels or a larger continuous range.

Each tree in the random forest algorithm is built on a resample of the original sample.  By convention, the random forest uses a bootstrap sample, with size equal to the original sample size, $n$. \cite{Strobl} found that using subsampling instead of bootstrap sampling to create individual trees within a random forest also helps to reduce variable selection bias. We hypothesize that these two bias sources may also cause biased estimation of $\hat{V}_{IJ}^B$ and explore variations on random forests that eliminate the variable selection bias to see if they perform well with the IJ variance estimator.

\subsection{Overview}

Although \cite{Wager} and \cite{Wager2} have proven that the IJ can be used to estimate the prediction variances of traditional random forests, their methods have not been tested using alternative individual tree types used in a random forest, such as conditional inference trees.  This variation is widely used and is important for eliminating variable selection bias. Here, we explore the applicability of the IJ to random forest variations, specifically using subsampling instead of bootstrap sampling and using CI trees instead of CART trees, and compare the accuracy of their estimates of prediction variances using simulation experiments.

\section{Methods}

\subsection{Data Simulation}

Ten different predictor variables ($X_1,...,X_{10}$) were generated by sampling from the normal distribution, with $X_1,...,X_5$  having mean zero and unit variance and $X_6,...,X_{10}$ having a mean of ten and variance of five. Eleven different simulation functions were then used to generate 11 different synthetic outcomes. Table \ref{table:simulation_functions} shows the name and corresponding simulation function used to generate each simulated dataset. Here, \textsc{AND} and \textsc{OR} are used to denote the unique characteristics of these simulated datasets derived from using the indicator function, $I(\cdot)$ (1 if the argument is true; 0 otherwise). Similarly, \textsc{SUM} and \textsc{SQ} are based on the summing and summing of the squares of the predictor variables, respectively. The number in each data simulation name corresponds to the number of predictor variables included in the simulation functions.  Note that less than the ten total predictor variables are used in each data simulation although all ten predictor variables are used in the construction of the random forests.  To simulate the data, 100 random test points were generated from each distribution and then 100 random training sets of varying size ($n$ = 200, 1000, 5000) were generated from each distribution. Eleven different distributions, each with three different sample sizes, resulted in 33 total types of simulated data sets.

\subsection{Random Forests}

All possible combinations of the proposed variations on random forests were implemented on the simulated data (both CI and CART trees, as well as bootstrap and subsampling).  For each variation, $m_{try}$ was set to three different default values: (1) $max\{floor(\frac{1}{3}p),1\}$, (2) $max\{floor(\frac{2}{3}p),1\}$, and (3) $max\{p,1\}$ (where $p$ is the total number of variables). For the current case of the simulations where $p = 10$, this resulted in $m_{try}$ sizes of 3, 6, and 10. Because $p = m_{try} = 10$ is the trivial case for bagged trees, we instead opted for an $m_{try}$ of 9. For subsampling random forest implementations, a subsample size of $n^{0.7}$ was used and all forests used $B=5n$ total trees, as recommended by \cite{Wager2}. Table \ref{table:n_s_B} shows the specific numbers for subsample and resample size for each corresponding total sample size. 

\subsection{Simulation Experiments}

\begin{figure}[!htbp]
	\centering
	\includegraphics{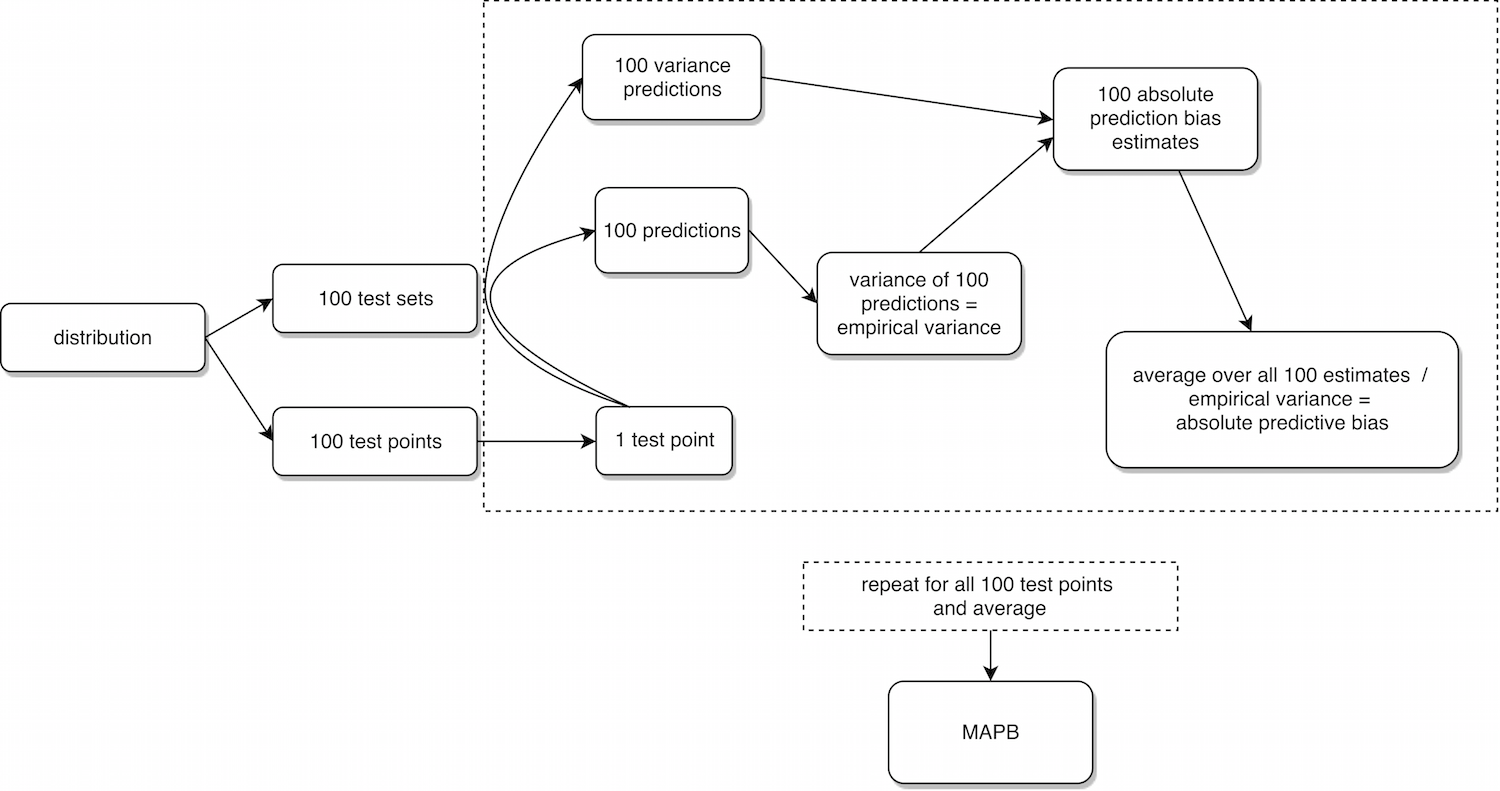}
	\caption{A diagram depicting the simulation experiments.  100 test points and 100 test tests were generated for each distribution and these were used to calculate the mean absolute predictive bias (MAPB).}
	\label{fig:sims_diagram}
\end{figure}

Figure \ref{fig:sims_diagram} contains a diagram depicting an overview of the simulation experiments.  Specifically, the four combinations of CI or CART trees with bootstrap or subsampling were implemented on all simulated data sets (Table \ref{table:simulation_functions}), each with sample sizes of 200, 1000, and 5000. $\hat{V}_{IJ}^B$ with the Monte Carlo correction as suggested by \cite{Wager} was calculated for all 100 test points, each using all 100 training samples.  The empirical prediction variance for each test point was calculated as the variance of all 100 predictions of each test point on the training samples. The absolute bias in $\hat{V}_{IJ}^B$ was calculated as the absolute difference of each variance estimate and the empirical prediction variance. The average absolute bias for each test point was calculated by averaging the absolute bias across all 100 data sets. The average absolute bias was normalized by the empirical variance and termed the ``absolute predictive bias'' in order to help interpretation and compare biases across different distributions and prediction ranges. Averaging the absolute predictive bias across all 100 test points resulted in the mean absolute predictive bias (MAPB) for each combination of tree type, resampling type, distribution, and sample size:
\begin{equation}
\label{eq:MAPB}
    MAPB = \frac{1}{100}\sum\limits_{k=1}^{100} \frac{\frac{1}{100}\sum\limits_{r=1}^{100}\vert\hat{V}_{IJ}\left(x^{(k)};Z^{(r)}\right) - \text{Var}_r\left[RF_s\left(x^{(k)};Z^{(r)}\right)\right]\vert}
    {\text{Var}_r\left[RF_s\left(x^{(k)};Z^{(r)}\right)\right]}
\end{equation}
where $k$ represents the index of each test point $x^{(k)}$, $r$ is the index of each training sample $Z^{(r)}$, and $RF_s\left(x^{(k)};Z^{(r)}\right)$ is the ensemble prediction of the $k^{th}$ test point using the $r^{th}$ training set.

\subsection{Statistical Computing}

All statistical computing was done in \texttt{R}, version 3.1.2 \citep{R}, using the \texttt{randomForest} \citep{randomForest} and \texttt{Party} packages \citep{hothorn2006unbiased} wrapped into the custom package, \texttt{RFinfer} \citep{cole_brokamp_2016_55336}. A software vignette, available online with the authors' software package, describes installation and usage examples for \texttt{RFinfer}.

\section{Results}

\subsection{Data Simulation}

Ten predictor variables were generated from the normal distribution in order to simulate the data.  $X_1,...,X_5$ were drawn from the standard normal distribution but $X_6,...,X_{10}$ were drawn from a normal distribution with a mean of ten and a variance of five. The last five predictor variables were created to have a larger range in order to observe the effects of the random forest variations compared to data with only small ranging predictors.  100 data sets were simulated for each of the 11 different simulation functions (Table \ref{table:simulation_functions}) and different sample sizes ($n$ = 200, 1000, 5000).  Furthermore, 100 test points used for prediction were generated for each simulation function. 

\subsection{Empirical Variance}

For each of the 100 test points, the variance of their predictions over all 100 test sets was calculated and termed the empirical variance. Table \ref{table-empvar} contains the median of these empirical variances for each distribution and sample size. As expected, the empirical variance increased within each type of distribution as more variables were used to generate the synthetic outcome and also decreased with increasing sample size. The $OR$ and $AND$ empirical variances were relatively small, all with a median less than 0.005.  This is likely because the distributions utilized the indicator function, reducing the synthetic outcome to only a few possible levels and defeating the effect of using predictors $X_6,...,X_{10}$, which had a larger range than $X_1,...,X_5$.  In contrast, the $SUM$ and $SQ$ distributions had a relatively larger empirical variance, especially $SUM5$ and $SQ5$, which utilized the predictors with a larger range and variation. 

\subsection{Bias in Variance Predictions}

\begin{figure}[!htbp]
	\centering
	\includegraphics{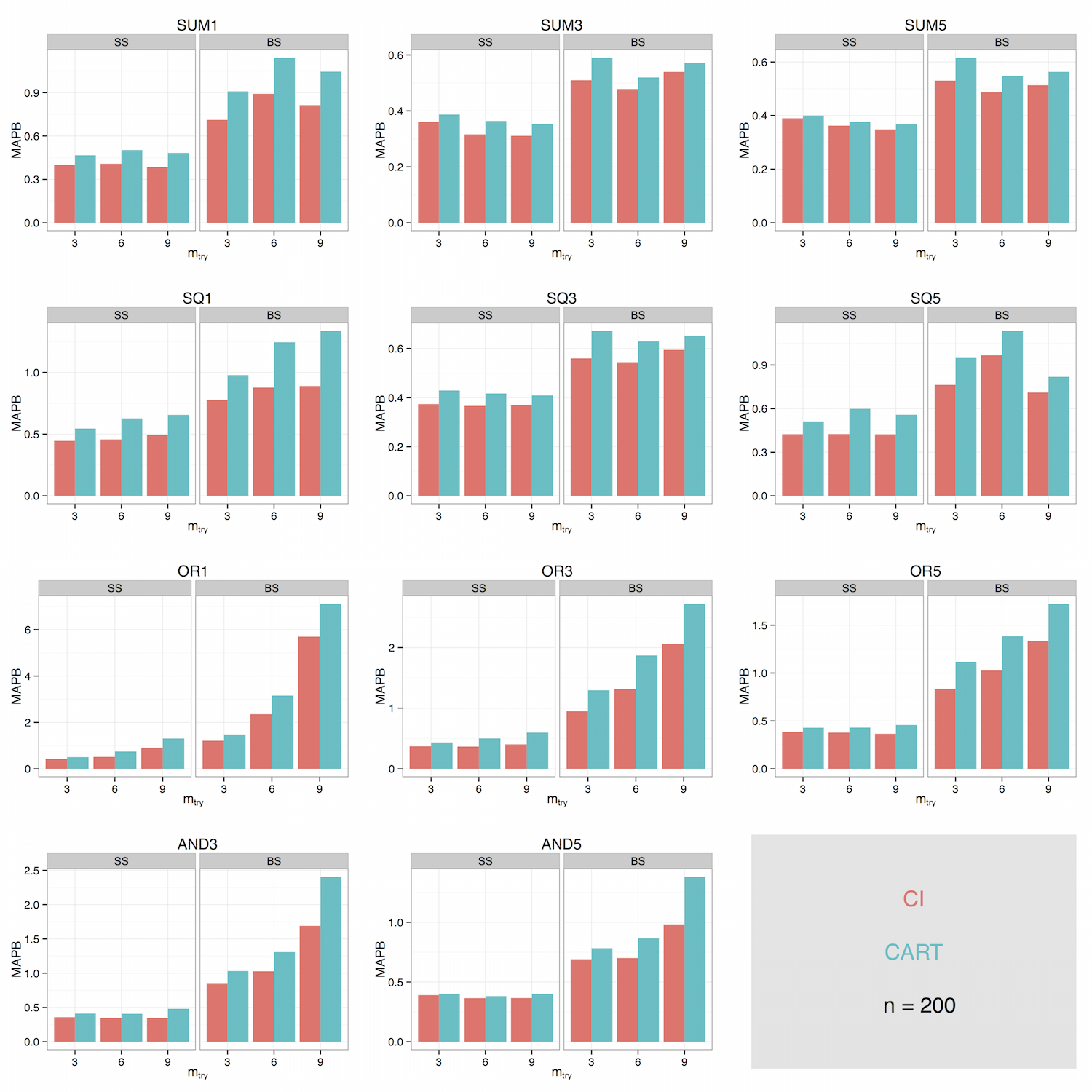}
	\caption{MAPB for each combination of subsample (SS) or bootstrap (BS) resampling, conditional inference (CI, red bars on left) or traditional CART (CART, green bars on right) trees, and $m_{try}$ for $n=200$.}
	\label{fig:results_n_200}
\end{figure}

\begin{figure}[!htbp]
	\centering
	\includegraphics{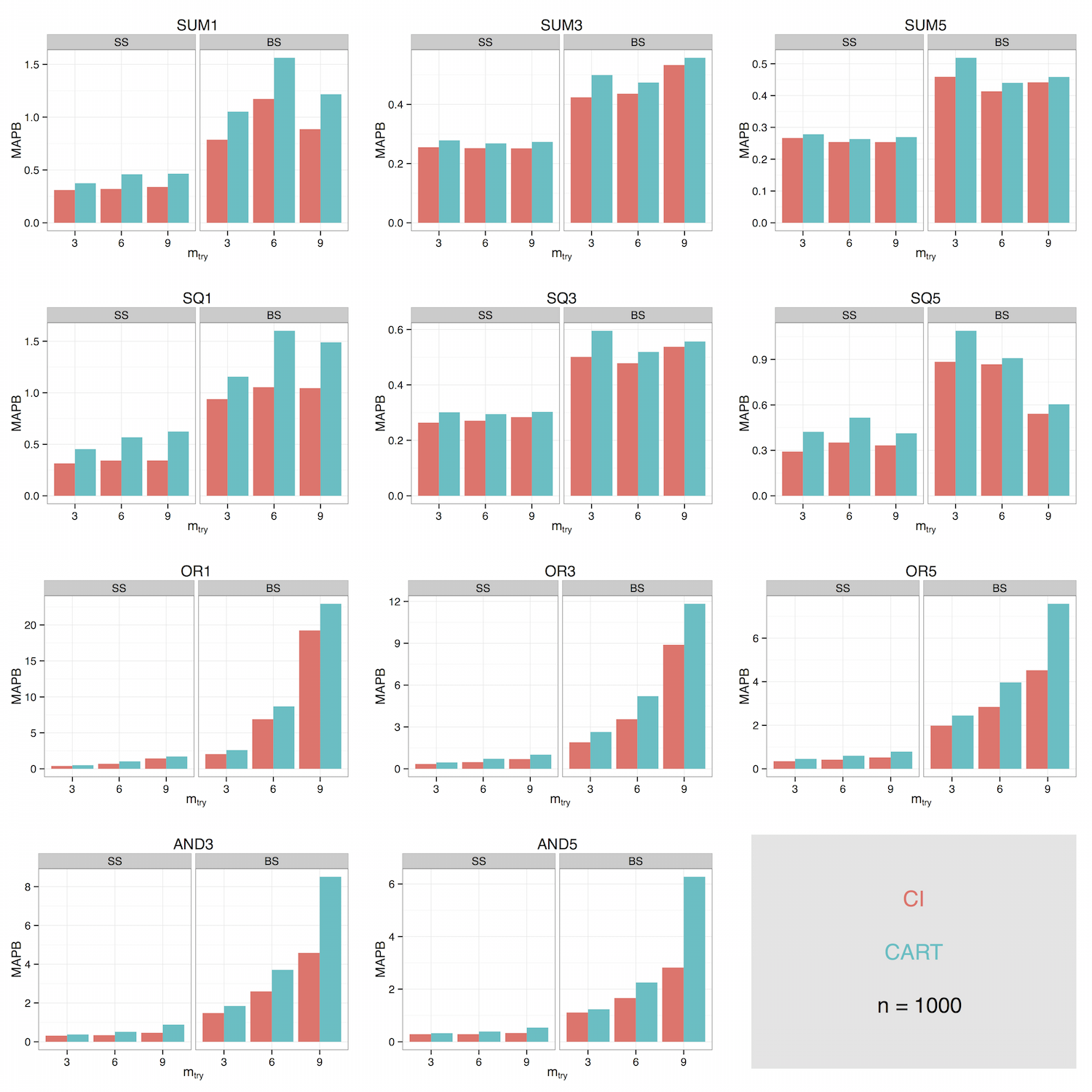}
	\caption{MAPB for each combination of subsample (SS) or bootstrap (BS) resampling, conditional inference (CI, red bars on left) or traditional CART (CART, green bars on right) trees, and $m_{try}$ for $n=1000$.}
	\label{fig:results_n_1000}
\end{figure}

\begin{figure}[!htbp]
	\centering
	\includegraphics{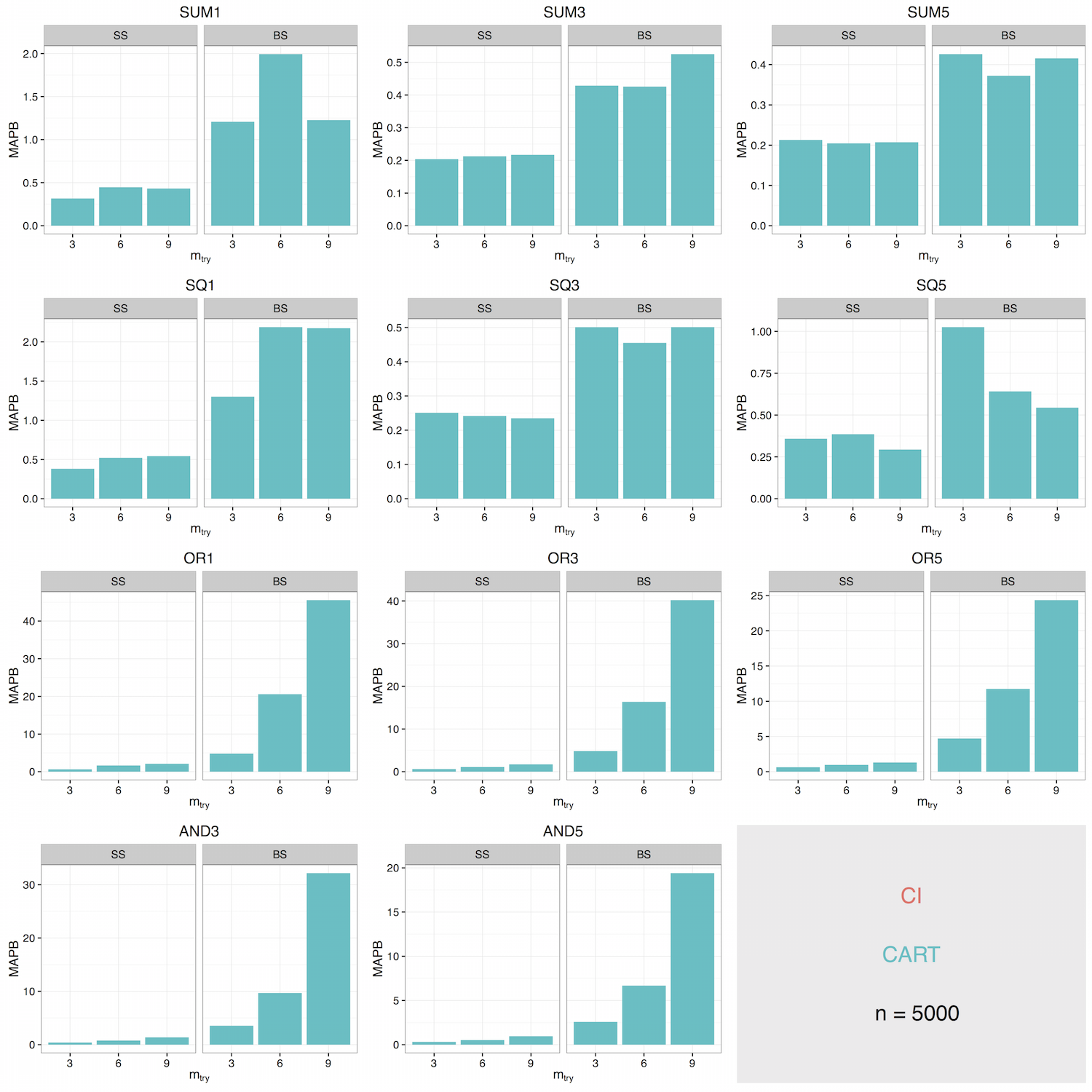}
	\caption{MAPB for each combination of subsample (SS) or bootstrap (BS) resampling, traditional CART (CART, green bars) trees, and $m_{try}$ for $n=5000$.}
	\label{fig:results_n_5000}
\end{figure}

The mean absolute predictive bias (MAPB) was calculated for each combination of resample type, tree type, $m_{try}$, sample size, and distribution by calculating the absolute difference in the variance estimate and the empirical variance, normalizing this bias by the empirical variance, and averaging over all 100 data sets and all 100 prediction points (Equation \ref{eq:MAPB}). See Figure \ref{fig:sims_diagram} for a diagram depicting the simulation experiments. The full results are presented in Table \ref{table:MAPB} and Figures \ref{fig:results_n_200}, \ref{fig:results_n_1000}, and \ref{fig:results_n_5000} show the MAPB for sample sizes of $n=200$, $n=1000$, and $n=5000$, respectively.  CI trees were not created for $n=5000$ due to computational limitations. Each of the simulation factors are explored in detail below.

\subsubsection{Tree Type}

The effect of tree type was consistent no matter the sample size, resampling method, or $m_{try}$ used; CI trees always had a lower MAPB than CART trees.  The decrease in MAPB when using CI trees instead of CART trees did not seem to differ with respect to sample distribution.

\subsubsection{Resampling Method}

The best improvement in MAPB resulted from utilizing subsampling rather than bootstrap sampling.  In fact, the worst performing simulation type using subsampling always performed better than the best simulation type using bootstrap sampling for every distribution. This was again the case for all combinations of sample size, $m_{try}$, and tree type. The difference was inflated when using a higher $m_{try}$ in the bootstrapped $OR$ and $AND$ distributions.

\subsubsection{Distribution}

Overall, the $SUM$ and $SQ$ distributions performed well, with MAPB of mostly less than one.  The $AND$ and $OR$ distributions, however, performed much worse, especially with increasing sample size and using bootstrap resampling and high $m_{try}$ values.

\subsubsection{\boldmath{$m_{try}$}}

Increasing $m_{try}$ caused a large increase in MAPB for the $OR$ and $AND$ distributions, but caused a smaller effect with varied directions on the $SUM$ and $SQ$ distributions. Using subsampling with the $OR$ and $AND$ datasets generally exhibited a small increase in MAPB with increasing $m_{try}$, whereas using bootstrap resampling with the $OR$ and $AND$ datasets exhibited a very large increase in MAPB with increasing $m_{try}$. Within the $SUM$ and $SQ$ distributions, $m_{try}$ had a much larger impact when bootstrap was used as the resampling method rather than subsampling. Here, $m_{try}$ had a varied effect when using bootstrap resampling that depended on the number of variables used in each distribution and the total sample size.

\subsubsection{Sample Size}

For the $SUM$ and $SQ$ distributions, increasing sample size decreased the MAPB for all combinations of $m_{try}$, tree type and resample type. However, the $OR$ and $AND$ distributions showed the opposite effect of increasing sample size, with a higher MAPB.  This effect was especially large with the bootstrap resampling method; for example, the MAPB of the random forests with an $m_{try}$ of nine trained on the $OR1$ distribution increased by an average of three fold when using $n=1000$ instead of $n=200$. 

\section{Discussion}

Here we have shown that using the IJ to estimate the variance of random forest predictions is much more accurate when using conditional inference trees instead of traditional CART trees and when using subsampling instead of bootstrap sampling. These simulation experiments show that the proofs in \cite{Wager2} hold when using CI trees instead of traditional CART trees under various simulated data sets of different distributions and sizes. 

Because there is no C implementation of the CI random forest method that indicates the number of times that each sample is included in each resample, the simulations using CI forests and a sample size of $n=5000$ were computationally infeasible. In order to carry out our simulations using $\hat{V}_{IJ}^B$, we had to use a pure R implementation of CI random forests.  This is different for CART random forests, where a C implementation already exists in the \texttt{randomForest} package.  However, it should be noted that the difference in computational times is due to the random forest creation step, not the implementation of $\hat{V}_{IJ}^B$.  This should not be an issue in the future when a C implementation of CI random forests is created.

The factor with the largest impact on MAPB was by far the resample method.  Implementing sub-bagging resulted in a more accurate estimation of the prediction variance, and this eclipsed the change in MAPB caused by any other variations in $m_{try}$, sample size or tree type. Although using CI trees was better than using CART trees, the performance increases was largest when using bootstrap resampling with the nonlinear $OR$ and $AND$ functions. However, the magnitude of improvement in MAPB was not increased for distributions utilizing the predictors with a wider range.  This result was not expected given the known bias of CART trees utilizing wider ranging predictors \citep{Strobl}.

The nonlinear distributions, $OR$ and $AND$ had an extremely small empirical variance compared to the $SUM$ and $SQ$ distributions.  Furthermore, the empirical variance decreased more rapidly with increasing sample size compared to the $SUM$ and $SQ$ distributions too. Thus, the increase in MAPB is likely due more to the decreased empirical variance rather then an increase in the absolute error of $\hat{V}_{IJ}^B$ and this is likely why the MAPB increased with increasing sample size for the $OR$ and $AND$ distributions, but decreased with increasing sample size for the $SUM$ and $SQ$ distributions.  

The key to the random forest model is decorrelation of the individual trees using $m_{try}$ and resampling.  Bootstrap resampling does decorrelate trees, with each resample showing an average number of distinct observations of $0.632n$ \citep{Hastie}. However, using subsampling with a subsample size of $n^{0.7}$ results in a far lower number of distinct original observations per resample (see Table \ref{table:n_s_B} for example).  Thus, subsampling creates more decorrelation in individual trees than bootstrap sampling and so $m_{try}$ makes a large difference in the performance of bootstrapped random forests because there is room for additional decorrelation, but not in subsampled random forests.  Similarly, the effects of $m_{try}$ are greater in the $AND$ and $OR$ distributions when using bootstrap resampling and not subsampling because the variance of the resamples are already so small that bootstrap resampling does not sufficiently decorrelate the individual trees, unlike subsampling. Overall, this is why the worst performing subsampled simulation still outperformed the best bootstrapped simulation.  Subsampling is likely more resistant to the correlation problems found in data with a lower variance and forests built with a higher $m_{try}$ value.  

Overall, our findings suggest that using CI trees instead of CART trees can be used in random forests to produce estimations of the prediction variance.  However, it is most important to use subsampling instead of bootstrap sampling as this has the largest impact on the accuracy of $\hat{V}_{IJ}^B$.

These simulations extend those performed previously by \cite{Wager2} by using different distributions, varying $m_{try}$ values, and including auxiliary noise variables in the training sets. However, in the future it would be valuable to evaluate the performance of $\hat{V}_{IJ}^B$ on correlated or multivariately distributed data, as well as on data with complex interactions because this is where random forest is most often used in real world settings.

\bibliographystyle{chicago}
\bibliography{rfci}

\clearpage

\section*{Tables}

\begin{table}[!htbp]
	\centering
	\caption{Ten functions used to simulate data.  $X_1,...,X_5$ were sampled from the normal distribution with mean 0 and variance 1. $X_6,...,X_{10}$ were sampled from the normal distribution with mean 10 and variance 5.}\label{table:simulation_functions}
	\begin{tabular}{cc}
		\hline
		\textbf{Name} & \textbf{Simulation Function} \\
		\hline
		\textsc{SUM1} & $X_1$ \\
		\textsc{SUM3} & $X_1 + X_3 + X_5$ \\
		\textsc{SUM5} & $X_1 + X_3 + X_5 + X_6 + X_7$ \\
		\textsc{SQ1} & $X_1^2$ \\
		\textsc{SQ3} & $X_1^2 + X_3^2 + X_5^2$ \\
		\textsc{SQ5} & $X_1^2 + X_3^2 + X_5^2 + X_6^2$ \\
		\textsc{OR1} & $I(X_1 > 0.4)$ \\
		\textsc{OR3} & $I(X_1 > 0.4) * I(X_3 > 0.6) * I(X_5 > 0.4)$ \\
		\textsc{OR5} & $I(X_1 > 0.4) * I(X_2 > 0.6) * I(X_3 > 0.4) * I(X_5 > 0.4) * I(X_6 > 6)$ \\
		\textsc{AND3} & $\frac{1}{3}[I(X_1 > 0.4) + I(X_2 > 0.6) + I(X_3 > 0.4)]$ \\
		\textsc{AND5} & $\frac{1}{5}[I(X_1 > 0.4) + I(X_3 > 0.6) + I(X_5 > 0.4) + I(X_6 > 6)]$ \\
		\hline
	\end{tabular}
\end{table}

\begin{table}[!htbp]
	\centering
	\caption{Three sample sizes ($n$) used for the simulated data sets and their corresponding subsample size ($s$), subsample fraction ($s/n$), and total resamples ($B$).}\label{table:n_s_B}
	\begin{tabular}{cccc}
		\hline
		$\boldsymbol{n}$ & $\boldsymbol{s=n^{0.7}}$ & $\boldsymbol{s/n}$ & $\boldsymbol{B=5n}$ \\
		\hline
		200 & 41 & 0.20 & 1,000  \\ 
		1,000 & 126 & 0.13 & 5,000 \\
		5,000 & 388 & 0.08 & 25,000 \\
		\hline
	\end{tabular}
\end{table}

\begin{table}[!htbp]
	\centering
	\caption[Empirical Variance]{Median of the empirical variances (Var) for each distribution and sample size.\label{table-empvar}}
	\resizebox{0.6\textwidth}{!}{%
		\begin{tabular}{c|c|c|c}
			\hline
			\textbf{Distribution} & \textbf{Var ($n=200$)} & \textbf{Var ($n=1000$)} & \textbf{Var ($n=5000$)}\\
			\hline
			SUM1 & 0.0055 & 0.0007 & 0.0001\\
			SUM3 & 0.0531 & 0.0192 & 0.0061\\
			SUM5 & 0.8661 & 0.2588 & 0.0946\\
			SQ1 & 0.0501 & 0.0088 & 0.0018\\
			SQ3 & 0.3512 & 0.1236 & 0.0404\\
			SQ5 & 78.8994 & 17.1922 & 6.1274\\
			OR1 & 0.0018 & 0.0004 & 0.0000\\
			OR3 & 0.0036 & 0.0007 & 0.0001\\
			OR5 & 0.0048 & 0.0009 & 0.0001\\
			AND3 & 0.0008 & 0.0001 & 0.0000\\
			AND5 & 0.0009 & 0.0002 & 0.0000\\
			\hline
		\end{tabular}
	}
\end{table}

\begin{table}[!htbp]
	\caption{The mean absolute predictive bias (MAPB) for each simulation, each the combination of a distribution, $m_{try}$, sample size, tree type, and resampling method.\label{table:MAPB}}
	\resizebox{1\textwidth}{!}{%
		\begin{tabular}{c@{\qquad}c@{\qquad}ccc@{\qquad}ccc@{\qquad}ccc@{\qquad}ccc}
			\toprule
			& & \multicolumn{6}{c}{\textbf{CART}} & \multicolumn{6}{c}{\textbf{CI}} \\
			\cmidrule{3-14}
			&  & \multicolumn{3}{c}{\textbf{Bootstrap}} & \multicolumn{3}{c}{\textbf{Subsample}} & \multicolumn{3}{c}{\textbf{Bootstrap}} & \multicolumn{3}{c}{\textbf{Subsample}} \\
			\cmidrule{3-14}
			\raisebox{4ex}[-8ex]{\textbf{Distribution}} & \raisebox{4ex}[-8ex]{$\boldsymbol{m_{try}}$} & $200$ & $1000$ & $5000$ & $200$ & $1000$ & $5000$ & $200$ & $1000$ & $5000$ & $200$ & $1000$ & $5000$ \\
			\cmidrule{3-14}
			& 3 & 0.91 & 1.05 & 1.21 & 0.47 & 0.37 & 0.32 & 0.71 & 0.79 &  & 0.40 & 0.31 &  \\ 
			& 6 & 1.14 & 1.56 & 1.99 & 0.50 & 0.46 & 0.45 & 0.89 & 1.17 &  & 0.41 & 0.32 &  \\ 
			\raisebox{2.5ex}[-5ex]{SUM1} & 9 & 1.04 & 1.22 & 1.23 & 0.48 & 0.46 & 0.43 & 0.81 & 0.89 &  & 0.39 & 0.34 &  \\ 
			& 3 & 0.59 & 0.50 & 0.43 & 0.39 & 0.28 & 0.20 & 0.51 & 0.42 &  & 0.36 & 0.26 &  \\ 
			& 6 & 0.52 & 0.47 & 0.43 & 0.36 & 0.27 & 0.21 & 0.48 & 0.44 &  & 0.32 & 0.25 &  \\ 
			\raisebox{2.5ex}[-5ex]{SUM3} & 9 & 0.57 & 0.56 & 0.53 & 0.35 & 0.27 & 0.22 & 0.54 & 0.53 &  & 0.31 & 0.25 &  \\ 
			& 3 & 0.62 & 0.52 & 0.43 & 0.40 & 0.28 & 0.21 & 0.53 & 0.46 &  & 0.39 & 0.27 &  \\  
			& 6 & 0.55 & 0.44 & 0.37 & 0.38 & 0.26 & 0.20 & 0.49 & 0.41 &  & 0.36 & 0.25 &  \\ 
			\raisebox{2.5ex}[-5ex]{SUM5} & 9 & 0.56 & 0.46 & 0.42 & 0.37 & 0.27 & 0.21 & 0.51 & 0.44 &  & 0.35 & 0.25 &  \\ 
			& 3 & 0.98 & 1.16 & 1.30 & 0.55 & 0.45 & 0.38 & 0.78 & 0.94 &  & 0.45 & 0.31 &  \\ 
			& 6 & 1.24 & 1.60 & 2.19 & 0.63 & 0.57 & 0.52 & 0.88 & 1.05 &  & 0.46 & 0.34 &  \\ 
			\raisebox{2.5ex}[-5ex]{SQ1} & 9 & 1.34 & 1.49 & 2.17 & 0.66 & 0.62 & 0.54 & 0.89 & 1.05 &  & 0.49 & 0.34 &  \\ 
			& 3 & 0.67 & 0.60 & 0.50 & 0.43 & 0.30 & 0.25 & 0.56 & 0.50 &  & 0.37 & 0.26 &  \\ 
			& 6 & 0.63 & 0.52 & 0.46 & 0.42 & 0.29 & 0.24 & 0.54 & 0.48 &  & 0.37 & 0.27 &  \\ 
			\raisebox{2.5ex}[-5ex]{SQ3} & 9 & 0.65 & 0.56 & 0.50 & 0.41 & 0.30 & 0.23 & 0.59 & 0.54 &  & 0.37 & 0.28 &  \\ 
			& 3 & 0.95 & 1.09 & 1.03 & 0.51 & 0.42 & 0.36 & 0.76 & 0.88 &  & 0.42 & 0.29 &  \\ 
			& 6 & 1.14 & 0.91 & 0.64 & 0.60 & 0.52 & 0.39 & 0.97 & 0.87 &  & 0.42 & 0.35 &  \\ 
			\raisebox{2.5ex}[-5ex]{SQ5} & 9 & 0.82 & 0.60 & 0.54 & 0.56 & 0.41 & 0.29 & 0.71 & 0.54 &  & 0.42 & 0.33 &  \\ 
			& 3 & 1.48 & 2.60 & 4.79 & 0.50 & 0.51 & 0.63 & 1.21 & 2.04 &  & 0.42 & 0.40 &  \\ 
			& 6 & 3.16 & 8.68 & 20.57 & 0.75 & 1.03 & 1.65 & 2.36 & 6.89 &  & 0.52 & 0.70 &  \\ 
			\raisebox{2.5ex}[-5ex]{OR1} & 9 & 7.12 & 22.94 & 45.56 & 1.31 & 1.71 & 2.09 & 5.70 & 19.24 &  & 0.91 & 1.44 &  \\
			& 3 & 1.30 & 2.64 & 4.82 & 0.44 & 0.46 & 0.61 & 0.95 & 1.90 &  & 0.37 & 0.35 &  \\ 
			& 6 & 1.87 & 5.21 & 16.36 & 0.50 & 0.72 & 1.10 & 1.31 & 3.56 &  & 0.37 & 0.48 &  \\  
			\raisebox{2.5ex}[-5ex]{OR3} & 9 & 2.72 & 11.83 & 40.21 & 0.60 & 1.01 & 1.72 & 2.06 & 8.89 &  & 0.40 & 0.70 &  \\
			& 3 & 1.12 & 2.45 & 4.71 & 0.43 & 0.46 & 0.63 & 0.84 & 1.98 &  & 0.38 & 0.35 &  \\   
			& 6 & 1.38 & 3.97 & 11.75 & 0.43 & 0.60 & 0.97 & 1.03 & 2.84 &  & 0.38 & 0.42 &  \\  
			\raisebox{2.5ex}[-5ex]{OR5} & 9 & 1.72 & 7.58 & 24.36 & 0.46 & 0.79 & 1.31 & 1.33 & 4.53 &  & 0.37 & 0.52 &  \\
			& 3 & 1.03 & 1.85 & 3.55 & 0.41 & 0.38 & 0.40 & 0.86 & 1.48 &  & 0.36 & 0.32 &  \\ 
			& 6 & 1.31 & 3.71 & 9.69 & 0.41 & 0.52 & 0.78 & 1.03 & 2.60 &  & 0.35 & 0.34 &  \\ 
			\raisebox{2.5ex}[-5ex]{AND3} & 9 & 2.41 & 8.51 & 32.16 & 0.48 & 0.89 & 1.37 & 1.69 & 4.58 &  & 0.35 & 0.46 &  \\ 
			& 3 & 0.78 & 1.24 & 2.57 & 0.40 & 0.33 & 0.32 & 0.69 & 1.11 &  & 0.39 & 0.29 &  \\ 
			& 6 & 0.87 & 2.25 & 6.68 & 0.38 & 0.39 & 0.52 & 0.70 & 1.66 &  & 0.37 & 0.29 &  \\ 
			\raisebox{2.5ex}[-5ex]{AND5} & 9 & 1.38 & 6.27 & 19.41 & 0.40 & 0.54 & 0.96 & 0.98 & 2.82 &  & 0.37 & 0.33 &  \\    
			\bottomrule
		\end{tabular}
	}
\end{table}

\clearpage

\section*{Figures}

\end{document}